# NEONATAL SEIZURE DETECTION USING CONVOLUTIONAL NEURAL NETWORKS

*Alison O'Shea, Gordon Lightbody, Geraldine Boylan, Andriy Temko*

Irish Centre for Fetal and Neonatal Translational Research, University College Cork

## ABSTRACT

This study presents a novel end-to-end architecture that learns hierarchical representations from raw EEG data using fully convolutional deep neural networks for the task of neonatal seizure detection. The deep neural network acts as both feature extractor and classifier, allowing for end-to-end optimization of the seizure detector. The designed system is evaluated on a large dataset of continuous unedited multi-channel neonatal EEG totaling 835 hours and comprising of 1389 seizures. The proposed deep architecture, with sample-level filters, achieves an accuracy that is comparable to the state-of-the-art SVM-based neonatal seizure detector, which operates on a set of carefully designed hand-crafted features. The fully convolutional architecture allows for the localization of EEG waveforms and patterns that result in high seizure probabilities for further clinical examination.

## 1. INTRODUCTION

Seizures in newborn babies are one of the most common indicators of a serious neurological condition, such as hypoxic-ischemic encephalopathy or stroke [1]. Most seizures in the neonatal stage occur at the subclinical level, meaning that they cannot be detected without monitoring the EEG [2]. Interpretation of neonatal EEG requires highly trained healthcare professionals and is limited to specialized units. This has prompted research into automated seizure detection systems where an objective score can be assigned to each segment of EEG [3].

When identifying a segment of EEG as seizure, a neurophysiologist is looking for a clear change in the EEG from its typical background behavior. In extracting a signature of the seizure signal with computer algorithms, data scientists usually follow one of two approaches – mimicking the clinician or using general data-driven EEG characteristics. On the assumption that rhythmic behavior is indicative of seizure, the former approach searches for the appearance of repetitive, pseudo-periodic waveforms by means of autocorrelation [4] or wave-sequence analysis [5]. The resultant characteristics are subjected to heuristic rules and compared with empirical thresholds [6], [7].

The data-driven approach involves extraction of features from time, frequency and information theory domains to capture energy, frequency, temporal and structural information to form a generic description of the EEG signal [8]. This approach involves extraction of non-stationary and model-based features and uses a classifier as a back-end [9]. The choice of the features for EEG representation and decision-making is prompted by an understanding that during a seizure the EEG epoch will become more ordered, more deterministic and quasi-periodic in nature, compared to the background EEG. With both approaches (data-driven and mimicking the clinician) the prior knowledge of neonatal EEG is reflected in a set of hand-crafted characteristics.

The Support Vector Machine (SVM) based algorithm, which is reported in [10], currently represents the state-of-the-art in the area of neonatal seizure detection. Its performance has been validated on a large clinical database to confirm robust functionality and a clinically acceptable level of accuracy of detection. This SVM system relies on a large set of engineered features.

Convolutional neural networks (CNNs) have achieved considerable success in the area of computer vision [11], [12]. The previously used manual feature-extraction approaches have been outclassed by deep CNNs, with their ability to learn relevant features from the data through end-to-end hierarchical representation and incorporation of translation invariance. Following the success in image processing, CNNs have been applied to speech, audio and music processing, where the time series can be represented as 'images' by means of time-frequency decomposition [13]–[15].

Similarly, EEG spectrogram images were used with CNNs in the prediction of epilepsy in adults [16], [17]. Likewise, EEG time-frequency images [18] or EEG spatial-temporal images [19], [20] were used together with CNNs in brain computer interfaces.

Unlike previous neonatal seizure detection studies, we aim to use CNNs to learn features from the raw EEG data, in a similar manner to CNNs that have been applied to raw audio [21]–[23]. The small (a few samples) width of the convolutional filters facilitates the learning of high frequency patterns in the first layers, and low frequency patterns, with the increased receptive field, in subsequent layers. The use of

The work was supported by Science Foundation Ireland Research Centre Award, INFANT (12/RC/2272). We gratefully acknowledge the support of NVIDIA Corporation with the donation of the TitanX GPU used for this research.

a fully convolutional neural network (FCNN), without any fully connected layers at the back-end, allows for the application of the developed seizure detector to a segment of EEG of any length. Additionally, the FCNN facilitates interpretation of the learnt feature maps. They provide the ability to localize the most discriminative waveforms in the original EEG, and offer higher representational power at lower computational cost.

FCNNs compute a deep nonlinear filter using only convolutional, pooling and activation function layers as components [24]. These building blocks all give translational invariance, meaning no computed parameters in the FCNN are location dependent. In this network a global average pooling (GAP) classification procedure replaces the fully connected layers which are commonly used as the classification layers in CNNs. The use of GAP and FCNN introduces regularization by reducing the number of learned parameters.

In this paper, a deep convolutional neural network is trained on raw neonatal EEG waveforms and its performance is compared with the state-of-the-art SVM-based classifier.

## 2. SVM-BASED BASELINE

A detector based on the SVM has previously been developed and validated on a clinical database [9, 10]. Its performance represents the state-of-the-art in neonatal seizure detection. SVMs are discriminative classifiers which are well suited to binary classification problems, such as the seizure detection task. Fig. **1** shows the system overview.

### 2.1. EEG pre-processing and segmentation

Raw EEG data, sampled at 256Hz, is filtered with a band pass filter with cut off frequency of 0.5 and 12.8Hz. The signal is then down-sampled to 32Hz. The EEG is then split into 8s epochs with 50% overlap.

### 2.2. Feature extraction

To learn an accurate representation of the EEG data, a set of 55 features are extracted from the preprocessed EEG; these come from the time, frequency and information theory domains to capture energy, frequency, temporal and structural information of an EEG epoch. The features are detailed in Table 1. The features are normalized prior to the classification stage to assure their commensurability.

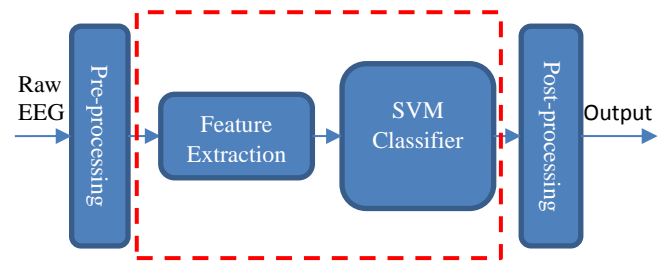

Fig. 1. Neonatal seizure detection system with feature extraction and SVM classifier.

Table 1. The SVM feature list is made up of 55 features which have been carefully engineered and optimized to give best performance on the neonatal seizure detection task.

| | Extracted Features |
|---|---|
| Frequency Domain | Total power (0-12Hz) |
| | Peak frequency of Spectrum |
| | Spectral edge frequency (80%, 90%, 95%) |
| | Power in 2HZ wide sub-bands |
| | Normalized power in sub-bands |
| | Wavelet energy |
| Time Domain | Curve length |
| | Number of maxima and minima |
| | Root mean squared amplitude |
| | Hjorth parameters |
| | Zero crossings (raw epoch, Δ, ΔΔ) |
| | Autoregressive modelling error (order 1-9) |
| | Skewness |
| | Kurtosis |
| | Nonlinear energy |
| | Variance (Δ, ΔΔ) |
| Information Theory | Shannon entropy |
| | Singular value decomposition entropy |
| | Fisher information |
| | Spectral entropy |

### 2.3. SVM training

A Gaussian kernel SVM was trained on the extracted features. Per-channel seizure annotations are required for training because neonatal seizures can be localized to a single EEG channel. Five-fold cross-validation was used on the training data to find the optimal set of hyper-parameters. Once the best set of hyper-parameters was chosen the model was trained using all of the training data. During testing, the classifier is applied separately to each EEG channel of the test subject. The test outputs from each of the 8 channels are converted to probabilistic values and fused during post-processing.

### 2.4. Post-processing

A moving average smoothing of 61 seconds is applied to the SVM output probabilities. The maximum probability across the channels is computed to represent the final probabilistic support for an epoch. The probability is then compared with a threshold and every positive decision is expanded by 30 seconds from each side, to account for the delay introduced by the smoothing operator.

## 3. A FCNN-BASED SEIZURE DETECTOR

A deep convolutional neural network is used here to replace the feature extraction and classification blocks in Fig. 1. The same pre-processing and post-processing routines as above are applied to the raw EEG data and the probabilistic outputs of the FCNN model, respectively.

### 3.1. Network architecture

The network is designed to be fully convolutional, without any fully connected layers, as illustrated in Table 2. The network layer values are calculated based on convolutional filters, including the final "decision making" layer at the output. The internal and final representations of the data within the network are based on local spatial filters, giving greater importance to positional relationships of features when making decisions. The FCNN architecture also means that the network can be used to process variable lengths of EEG input data (which is not yet exploited in this study).

The convolutional layers in the network use 1-dimensional convolutional filters of length 4, constructed using the Keras deep learning library [25]. There are 6 convolutional layers in total and each can be seen as a higher level of feature abstraction of the previous layer.

The output of each feature map, for the rectified linear unit non-linearity, is calculated as

$$h_i^k = \max(0, ((W^k * x)_i + b_k)). \qquad (1)$$

Here, $i$ is the temporal index within a feature map, $x$ is a matrix representing the input to a layer. $(W^k * x)_i$ refers to convolutional calculation, at position $i$, for the $k^{th}$ feature map, taking inputs from all feature maps in the previous layer. Each convolutional layer weight matrix, $W$, is a 3D tensor, containing elements for every combination of the destination feature map, source feature map and source position. The rectified linear unit activation is applied by taking the max between the output and 0.

The same epoch length (8 seconds), as with the SVM-based systems, is used as an input to FCNN. The main difference is the fact that the input is raw EEG, not the extracted features. Each input window is shifted by 1 second, which can be seen as a data augmentation step.

Pooling reduces the dimensionality of the data, and thus the number of parameters in the network, and provides spatial invariance. The window used for pooling in this study are of width 8 with a stride of 2, halving the size of the input. We used *average* pooling. Convolution followed by average pooling can be seen as a filter-bank signal decomposition performed on the raw EEG signal. We have also tried *max* pooling, which led to slightly inferior results in our experiments. The *max* pooling operator intuitively implies that instead of quantifying a particular waveform, which are detected by a matched filter in the convolutional layers, we are rather looking for a presence of a particular waveform.

In the final convolutional layer two feature maps represent the input data which maintain the original temporal ordering. The final convolutional output is followed by GAP giving 2 values on which the softmax operation is performed to give the seizure and non-seizure class probabilities. It is common to use a fully connected layer as a backend in CNNs, which ultimately makes the classification based on the CNN extracted features. By using GAP the two final feature maps are forced to correspond to the seizure or non-seizure class, which allows for a clearer visualization and localization of the underlying processes, as each sample in the final layer can be traced back to a corresponding window of the input EEG signal. The first layer filters have a receptive field of 4 samples whereas the final layer filters have a receptive field of 47 input samples which corresponds to approximately 1.5 seconds of raw EEG (at the 32Hz sampling rate). It is thus possible to conclude that high pass filters are learnt in the first layers, which have narrow receptive fields, and low pass filters are learnt in the final layer, with the hierarchically increased width of receptive field.

Table 2. CNN architecture, all convolutional layers are followed by rectified linear units.

| Layer Type | Shape | Output Shape | Parameters |
|---|---|---|---|
| Input | 256 | 256x1 | 0 |
| 1D Convolution | 32 filters 4x1 kernel Stride 1 | 32x253x1 | 160 |
| 1DConvolution | 32 filters 4x1 kernel Stride 1 | 32x250x1 | 4128 |
| 1DConvolution | 32 filters 4x1 kernel Stride 1 | 32x247x1 | 4128 |
| Batch Norm. | | 32x247x1 | 64 |
| Average Pooling | Pool 8 Stride 2 | 32x120x1 | 0 |
| 1DConvolution | 32 filters 4x1 kernel Stride 1 | 32x117x1 | 4128 |
| 1DConvolution | 32 filters 4x1 kernel Stride 1 | 32x114x1 | 4128 |
| Average Pooling | Pool 4 Stride 2 | 32x56x1 | 0 |
| 1DConvolution | 2 filters 4x1 kernel Stride 1 | 2x53x1 | 258 |
| GAP | | 2x1 | 0 |
| Softmax | | 2 | 0 |

### 3.2. Training parameters and optimization

The network was trained using categorical cross-entropy as the loss function. Stochastic gradient descent was used with an initial learning rate of 0.003, this was reduced by 10% every 20 iterations. Nesterov momentum was set to 0.9. A batch size of 2048 was used for training and validation.

For regularization, batch normalization was used [26] to normalize the intermediate layer outputs, which speeds up the learning process. By reducing the internal covariate shift, internal states appear normalized when presented to deeper layers, allowing for the use of higher learning rates, this largely removes the need for dropout. CNNs also have built-in regularization, using shared weights and sparse connectivity results in fewer trainable parameters than fully connected networks; this also reduces the risk of overfitting on training data. For this reason, no early stopping criteria was found necessary in this study. It was found that the network converges to a stable validation accuracy after 60 learning iterations. An example of training procedure is shown in Fig. 2. The training and validation performances are computed without post-processing and thus are lower than the testing data performance. It can be seen that the validation performance stabilizes at around 60 iterations. Curves shown were generated using patient 9 as the test case.

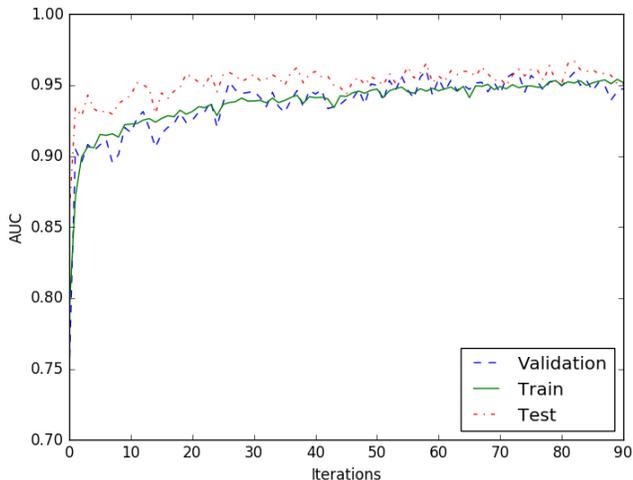

Fig. 2. The evolution of performance on training, validation and test data for one subject against training iterations.

## 4. RESULTS AND DISCUSSION

### 4.1 Database

The dataset used in these experiments contains EEG recorded in the Neonatal Intensive Care Unit of Cork University (NICU) Maternity Hospital, Ireland. It includes recordings from 18 term newborns that experienced seizures from hypoxic ischemic encephalopathy brain injury. The EEG dataset totals 835 hours in duration and comprises of 1389 seizures. Eight channels of EEG were recorded at 256Hz using the 10-20 placement system modified for neonates, using the following 8 channels in bipolar montage F4-C4, C4-O2, F3-C3, C3-O1, T4-C4, C4-Cz, Cz-C3, C3-T3. All seizure events in the recordings were annotated by two independent neonatal electro-encephalographers. No data pre-selection or removal of artifacts was performed, and the data is reflective of a real-world NICU environment. The same dataset has been used in [10] and the performances are directly comparable.

### 4.2. Performance assessment and metrics

The area under the receiver operating characteristic curve (AUC) was used as the primary measure of the classifier performance. The ROC curve is a plot of sensitivity versus specificity, where sensitivity corresponds to the percentage of correctly identified input epochs containing seizures and specificity corresponds to the accuracy of identifying non-seizure epochs. For a clinically acceptable performance levels, high specificity is a requirement which implies that less false alarms will be generated. For this reason the AUC where the specificity is higher than 90% is also reported as in [10].

Table 3. A comparison of FCNN and SVM results.

| Patient | AUC (%) | | AUC90 (%) | |
|---|---|---|---|---|
| | SVM | CNN | SVM | CNN |
| 1 | 95.5 | **96.9** | 83.1 | **87.1** |
| 2 | 99.4 | 99.4 | 93.5 | **94.1** |
| 3 | 97.2 | **97.3** | **81.6** | 80.0 |
| 4 | **97.2** | 96.0 | **82.6** | 76.9 |
| 5 | 94.7 | **98.1** | 69.8 | **86.3** |
| 6 | **96.0** | 95.6 | **73.0** | 67.2 |
| 7 | **99.1** | 98.7 | **91.5** | 87.5 |
| 8 | 97.7 | **98.0** | 81.0 | **85.4** |
| 9 | **99.0** | 98.9 | **91.8** | 91.7 |
| 10 | 88.6 | **97.2** | 61.7 | **85.3** |
| 11 | 97.6 | **98.1** | **88.6** | 87.9 |
| 12 | 95.3 | **96.1** | **76.4** | 75.7 |
| 13 | **98.7** | 95.3 | **89.0** | 70.6 |
| 14 | 95.9 | **97.6** | 85.5 | **88.5** |
| 15 | 97.8 | 97.8 | 87.4 | **88.8** |
| 16 | 93.1 | **93.9** | 73.8 | **74.6** |
| 17 | 96.6 | **98.2** | 91.3 | **95.7** |
| 18 | **97.8** | 93.8 | **85.3** | 68.1 |
| Average | 96.5 | **97.1** | 82.6 | **82.9** |

In a clinical setting, the EEG of a neonate in the NICU cannot be seen before the baby is born. The classifier thus needs to be assessed in a patient independent way. The leave-one-patient-out cross-validation (LOO) method is the most suitable test procedure to test a patient independent seizure detector. In this method, data from 17 newborns is used to train the classifier, and the performance is then assessed on the remaining newborn's data. This routine is repeated 18 times, using each baby as the unseen test case.

### 4.3. Comparison of seizure detection performance

Table 3 shows the AUC and AUC90 results for the SVM and FCNN systems. The FCNN results in the AUC of 97.1% and AUC90 of 82.9%. This marginally outperforms the results of the SVM system on both metrics. Observing the AUC values, FCNN scores better in 10 out of the 18 patients, with the SVM scoring better in 6, and equal performance obtained on 2 patients. These results show that without any feature engineering, selection, or extraction the proposed FCNN architecture manages to achieve comparable results, by end-to-end learning of discriminative patterns, to differentiate between the background EEG and seizures. The network extracts meaningful representations of the data from sample level inputs, and uses these to create high level features in the deeper network layers.

In terms of the number of parameters, the FCNN requires 16,930 parameters as shown in Table 2. The SVM classification system alone from [10] consists of ~10,000 support vectors of 55 dimension each, which totals over 560,000 parameters (includes the Lagrangian multipliers). This number does not include the parameters that have to be tuned for each feature during feature extraction, kernel parameters, etc. It can be seen that the comparable performance with FCNN is achieved with a much smaller number of parameters, and without extensive work on the feature engineering side.

Interestingly, the FCNN deep architecture is more interpretable than the shallow SVM-based system. The features extracted by the FCNN can be visualized [27], and may also contribute to the overall understanding of EEG signals and neonatal brain functioning.

It is worth discussing the various DNN architectures that were tried on the way to developing the one presented in this paper. Initially, the EEG was converted to time-frequency images (spectrograms) and 2D CNNs were utilized, adopted from the area of image processing [17] – this architecture proved unsuccessful in the seizure detection task. The problem was then reformulated as one of learning the filter-bank weights of the spectral envelope; this is similar to mel-scaling which is used in speech processing [12]. Some improvement was obtained with respect to the previous experiment, but the performance was still inferior to that of the SVM baseline.

The next stage was to use 1D CNNs applied to the raw EEG waveform [23]. EEG cannot be considered stationary even on a short-time scale, e.g. because of the presence of spikes and other discrete events; thus learning data-driven filters instead of using the cos/sin basis of the FFT is sensible. This led to 1D CNNs with wide convolutional filters (1-4s, 32-128 samples) which significantly improved the performance, approaching that of the SVM baseline. Finally, sample size filters were used, which in contrast to larger filter lengths allow for the learning of the various filters in a hierarchical manner [21]. In this paper filters of length 4 samples, corresponding to 0.125s, were used.

### 4.4. Examples of FCNN-based detections

The temporal relationship between the final layer and the raw input EEG allows for replication of the process by which a healthcare professional interprets the EEG. In all layers of the network, temporal ordering is maintained, any spikes of seizure or non-seizure activity in each of these feature maps can be traced back to the corresponding receptive field in the raw input EEG. This is not feasible with shallow architectures built on heavily engineered features – the features come from different domain and it is hardly possible to trace-back and find out to which feature a particular (nonlinear) decision was mostly attributable to. In contrast, the two final feature maps in the last convolutional layer correspond to the seizure and non-seizure cases. The amplitudes in this layer are indicative of the extent to which a particular EEG waveform resembles seizure or non-seizure. By viewing these waveforms we can learn about what the network interprets as being representative of seizures in the input signal.

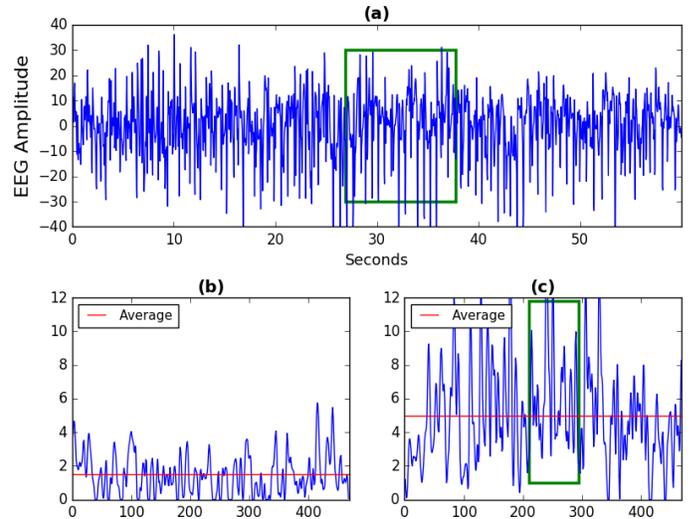

Fig. 3. One minute of EEG annotated as being seizure activity by an expert. The raw EEG input is shown in (a). The values of each feature map in layer 6 of the CNN are also shown in (b) and (c).

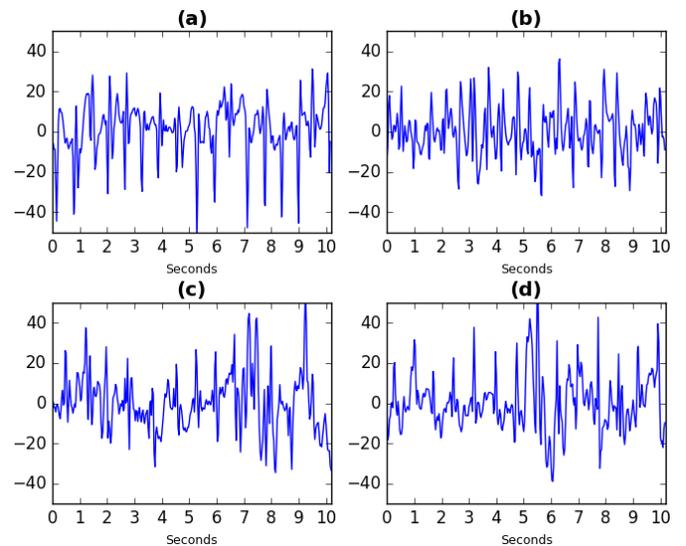

Fig. 4. Four examples of EEG with high seizure probability, as calculated by the FCNN.

Fig. 3 (a) shows one minute of EEG seizure activity in the time domain from a patient which was excluded from the training data. The network classified this segment as seizure, with a probability of 0.97. The two feature maps in the final network layer, before GAP, can be seen in Fig. 3, where feature map (c) represents the seizure class. Using this feature map a segment in the time domain that corresponds to high values of the seizure feature map can be identified as highlighted in green in Fig. 3.

Fig. 4 displays four examples of raw EEG which the network considers to be highly representative of seizure activity. All of these samples were also judged to represent seizures by clinical experts. Fig. 4 (a) plots the 10 second clip highlighted in Fig. 3. Fig. 4 (b) is taken from the same patient whereas plots (c) and (d) are taken from a different patient.

## 5. CONCLUSION

This work presented how fully convolutional neural network architecture with sample-size filters could be applied to raw EEG to learn relevant features and classify segments of neonatal EEG into either seizure or non-seizure classes. Evaluated on a large dataset of unedited EEG, the developed novel system has achieved a level of performance which is comparable to the previously developed SVM-based system which was built on top of a large set of heavily engineered features. The advantages of the proposed approach, such as the end-to-end optimization of feature extraction and classification, have been discussed.

The main advantage of using a fully convolutional network is that the weight matrix at each convolutional layer can be visualized, and so further research will concentrate on gaining new insights into the make-up of neonatal EEG signals based on these learned weight matrices.